\documentclass[lettersize,journal]{IEEEtran}
\usepackage{amsmath,amsfonts}
\usepackage{algorithmic}
\usepackage{algorithm}
\usepackage{array}
\usepackage[caption=false,font=normalsize,labelfont=sf,textfont=sf]{subfig}
\usepackage{textcomp}
\usepackage{stfloats}
\usepackage{url}
\usepackage{verbatim}
\usepackage{graphicx}
\usepackage{cite}
\usepackage{booktabs}
\usepackage{bbding}
\usepackage{threeparttable}
\usepackage{multirow}
\usepackage{orcidlink}

 \usepackage{amsmath} 
 \usepackage{amssymb}
 \usepackage{adjustbox}
 \usepackage{pifont}
 \usepackage{makecell}
 \newcommand{\cmark}{\ding{51}} 
 \newcommand{\xmark}{\ding{55}}
 \usepackage{times}
\usepackage{helvet}
\usepackage{courier}
\usepackage{xcolor}

\usepackage{xurl}

\begin{document}

\title{RIS-LAD: A Benchmark and Model for Referring Low-Altitude Drone Image Segmentation}

\author{Kai Ye \orcidlink{0000-0003-0914-0213}, Yingshi Luan, Zhudi Chen, Guangyue Meng, Pingyang Dai \orcidlink{0000-0001-9780-271X}, Liujuan Cao \orcidlink{0000-0002-7645-9606}, \textit{Member, IEEE}
\thanks{ 
Kai Ye, Yingshi Luan, Zhudi Chen and Guangyue Meng are with the Key Laboratory of Multimedia Trusted Perception
 and Efficient Computing, Ministry of Education of China, and the Institute
 of Artificial Intelligence, Xiamen University, Xiamen 361005, China (e-mail:
 yekai@stu.xmu.edu.cn; 37220232203770@stu.xmu.edu.cn; zhudichen@stu.xmu.edu.cn; mgy1881@stu.ouc.edu.cn).

  Liujuan Cao \textit{(Corresponding author)} and Pingyang Dai are with the Key Laboratory of Multimedia Trusted Perception and Efficient Computing, Ministry of Education
 of China, Xiamen University, Xiamen 361005, China, and also with the
 School of Informatics, Xiamen University, Xiamen 361005, China (e-mail:
 caoliujuan@xmu.edu.cn, pydai@xmu.edu.cn).
 }}



\maketitle

\begin{abstract}
Referring Image Segmentation (RIS), which aims to segment specific objects based on natural language descriptions, plays an essential role in vision-language understanding.
Despite its progress in remote sensing applications, RIS under Low-Altitude Drone (LAD) scenarios remains underexplored, as existing datasets and methods are typically designed for high-altitude and static-view imagery. They struggled to handle the unique characteristics of LAD views, such as diverse viewpoints and high object density.
In this paper, we propose RIS-LAD, the first fine-grained RIS benchmark tailored for LAD scenarios, featuring 13,871 meticulously annotated image-text-mask triplets collected from real-world drone footage with emphasis on small, densely cluttered objects and multi-view perspectives.
It highlights new challenges not present in previous benchmarks, such as category drift caused by tiny objects and object drift under crowded objects of the same class. 
Additionally, we propose the Semantic-Aware Adaptive Reasoning Network (SAARN), which decomposes and adaptively routes semantic information to different network stages rather than uniformly injecting all linguistic features.
Specifically, the Category-Dominated Linguistic Enhancement (CDLE) aligns visual features with object categories during early encoding, while the Adaptive Reasoning Fusion Module (ARFM) dynamically selects semantic cues across scales to enhance reasoning in complex scenes. Extensive experiments reveal that RIS-LAD presents substantial challenges to state-of-the-art RIS algorithms, and also demonstrate the effectiveness of our proposed model in addressing these challenges.
The dataset and code will be publicly released soon at: \url{https://github.com/AHideoKuzeA/RIS-LAD-A-Benchmark-and-Model-for-Referring-Low-Altitude-Drone-Image-Segmentation}.
\end{abstract}


\section{Introduction}
\IEEEPARstart{L}, which typically operate below 200 meters, have become widely utilized in real-world applications due to their flexible deployment and high versatility \cite{banafaa2024comprehensive,casanova2025comparison,li2025cooperative}.
This trend has sparked increasing research interest in vision tasks under LAD scenarios, leading to the development of benchmarks such as VisDrone~\cite{visdrone} and UAV123~\cite{UAV123}, which support object detection, tracking and other related tasks~\cite{DroneCrowd,Okutama-Action,UAVDT}.
Recently, visual understanding in LAD scenarios~\cite{sun2025refdrone,li2025aeroreformer} has attracted growing interest, especially in multi-modal tasks.
Among them, Referring Image Segmentation (RIS) \cite{liu2025hybrid,wang2025iterprime,huang2025densely} is a fundamental perception task that aims to segment objects from images based on descriptions.
Incorporating RIS into LAD systems enables LADs to better accommodate a wider range of practical applications.

\begin{figure}
    \centering
    \includegraphics[width= \linewidth]{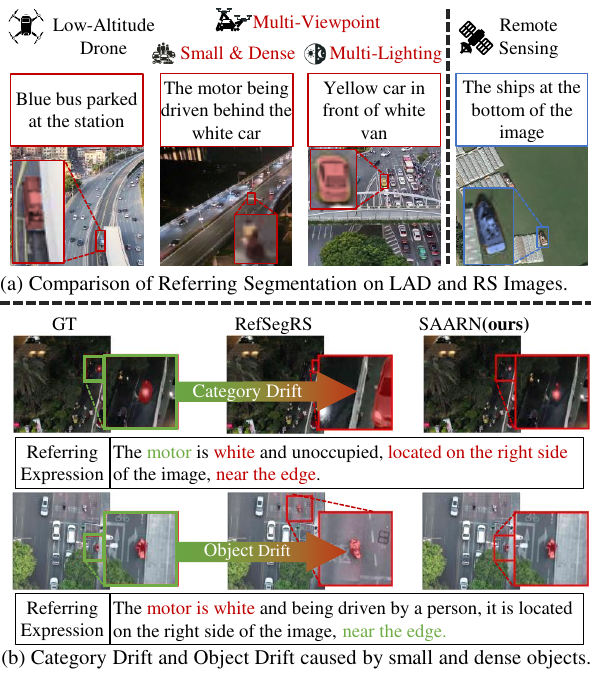}
    \caption{Illustration of the challenges in referring low-altitude drone image segmentation (RLADIS). The method shown in (b) is RefSegRS \cite{chen2025rsrefseg}, one of the current SOTA approaches for RRSIS.}
    \label{first}
\end{figure}

However, existing RIS research mainly focuses on conventional scenes or high-altitude remote sensing imagery, leaving the distinctive properties of LAD imagery largely unexplored in the current literature.
Although previous Referring Remote Sensing Image Segmentation (RRSIS) studies~\cite{shi2025multimodal,chen2025rsrefseg,pan2024rethinking} have provided valuable insights, they are mostly based on imagery taken from satellites or helicopters.
As shown in Fig.~\ref{first}(a), Referring Low-Altitude Drone Image Segmentation (RLADIS) presents several unique challenges compared to RRSIS: (1) diverse viewpoints due to lower and more flexible camera positions; (2) diverse illumination conditions; and (3) smaller and more densely distributed objects.
These challenges introduce a substantial domain gap that hinders the effective generalization of current RRSIS methods to RLADIS tasks.

To bridge this gap, we propose RIS-LAD, the first dataset designed for fine-grained RLADIS, accompanied by a semi-automatic annotation pipeline that ensures high-quality referring expression generation and annotation efficiency.
The final dataset contains 13,871 image-text-mask triples, providing a benchmark for evaluating RLADIS methods.
Furthermore, our analysis reveals two key challenges that hinder the direct adaptation of RRSIS methods to RLADIS: \textbf{category drift} and \textbf{object drift}.
As shown in Fig.~\ref{first}(b), category drift arises when the object occupies only a small region of the image, misleading the model to focus on larger, semantically similar objects.
Object drift, on the other hand, results from a high density of instances belonging to the same category, making it difficult for the model to correctly identify the specific object referenced in the expression.

Motivated by these drift issues, we propose the Semantic-Aware Adaptive Reasoning Network (SAARN), which decouples linguistic features and injects them at appropriate stages of the network to ensure semantic alignment.
Unlike existing decoupling-based approaches that uniformly inject all linguistic components across modules \cite{lei2024exploring,zhang2025referring}, SAARN employs a more targeted strategy. 
Specifically, the Category-Dominated Linguistic Enhancement (CDLE) module injects only class-level linguistic features into the encoder. This injection aligns early visual representations with accurate category-level semantics. 
Meanwhile, global linguistic features are selectively integrated via class-guided gating, reinforcing semantic consistency.
In the multi-scale fusion stage, the Adaptive Reasoning Fusion Module (ARFM) performs scale-aware enhancement by dynamically weighting semantic cues across multi-scale features.
This mechanism simulates a reasoning process, guiding the model to infer the most semantically consistent instance among densely packed objects of the same category.

The main contributions of this paper are as follows: 

\begin{itemize}
    \item We propose RIS-LAD, the first fine-grained referring image segmentation dataset tailored for low-altitude drone scenarios. 
    It offers diverse LAD scenes with rich textual descriptions and high-quality segmentation masks.
    \item To address category drift and object drift in RLADIS, we propose the Semantic-Aware Adaptive Reasoning Network (SAARN). This model focuses on aligning semantic categories and selectively integrates the most relevant linguistic features at optimal feature scales.
    \item We conducted extensive experiments to investigate the challenges inherent in RLADIS and construct a dedicated benchmark for this task. Our proposed SAARN achieves state-of-the-art performance on the RIS-LAD benchmark.
\end{itemize}

\section{RELATED WORKS}
\subsubsection{Referring Image Segmentation (RIS)}aims to localize target objects at the pixel level in an image based on natural language expressions \cite{hu2016segmentation}. 
A representative line of methods focuses primarily on enhancing vision-language alignment to improve segmentation performance \cite{chng2024mask,wang2024unveiling,shah2024lqmformer}. 
VATEX \cite{nguyen2025vision} leverages a CLIP-based prior module to generate heatmaps, which enforces semantic consistency and improves contextual alignment between language and vision.
ASDA \cite{yue2024adaptive} employs a dual-alignment mechanism with dynamic feature selection to better align visual and linguistic modalities.
Other approaches explore novel architectural designs for RIS \cite{yang2024remamber,chen2024sam4mllm,Xia_2024_CVPR}.
IteRPrimE \cite{wang2025iterprime} refines visual activation maps through an iterative Grad-CAM refinement strategy and adopts a primary-word emphasis module to improve the identification of key semantic components in language expressions.
Although these approaches demonstrate promising performance on conventional imagery, their effectiveness deteriorates notably when applied to LAD imagery.

\subsubsection{Referring Remote Sensing Image Segmentation (RRSIS)} is a domain-specific extension of the RIS task for remote sensing images, first introduced by \cite{sun2022visual}.
Given the high resolution and small object sizes in remote sensing images, many recent approaches have adopted multi-scale feature fusion \cite{ma2025lscf,lu2025rrsecs}.
FIANet \cite{lei2024exploring} disentangles various components of the referring expression and incorporates a text-aware multi-scale enhancement module to improve multi-modal discrimination.
RMSIN \cite{liu2024rotated} designs rotation-aware modules to precisely segment objects based on the unique properties of remote sensing imagery.
Meanwhile, pre-trained large vision models have drawn growing interest in this field \cite{dong2025diffris}.
RSRefSeg integrates a CLIP-based \cite{radford2021learning} module to capture implicit visual activations and uses them as guidance prompts for SAM \cite{kirillov2023segment}.
For benchmarks, existing RRSIS datasets  \cite{dong2024cross,liu2024rotated,yuan2024rrsis} are mainly constructed from high-altitude sources such as Google Earth and GF-2, which feature scenes with larger objects and fixed viewpoints. 
Although some studies have explored low-altitude scenarios \cite{sun2025refdrone}, challenges such as tiny objects and high instance density persist due to diverse viewpoints. Existing LAD-based datasets like UAVid-RIS and VDD-RIS \cite{li2025aeroreformer} remain limited in object categories and expression granularity, restricting their ability to support fine-grained RIS tasks.

\begin{figure}
    \centering
    \includegraphics[width=0.9\linewidth]{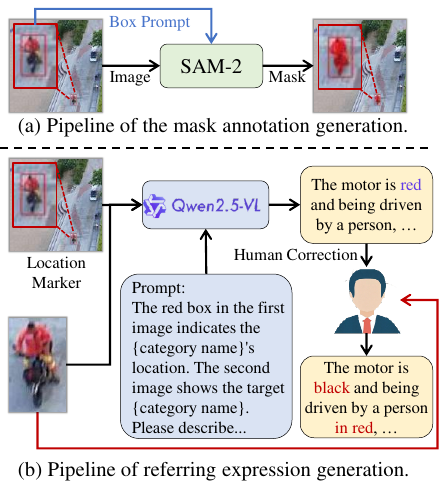}
    \caption{Illustration of the semi-automatic annotation pipeline for RIS-LAD. Mask generation leverages SAM-2 to extract instance-level segmentation masks. Referring expression generation employs Qwen2.5-VL to produce initial descriptions conditioned on location markers and cropped object instances, with subsequent human correction.}
    \label{annotation pipeline}
\end{figure}

\section{RIS-LAD Dataset}
This section introduces the construction process and key characteristics of the RIS-LAD. Full dataset details and ethical statements are provided in the appendix.
\subsubsection{Image and Category Selection} 
The suitable images for the RIS task are selected from the publicly available CODrone dataset \cite{ye2025clearflexibleprecisecomprehensive}, which is designed for LAD-oriented object detection.
The selected images encompass diverse real-world urban scenes under various lighting conditions and from different viewpoints.
Following common procedures for constructing RRSIS datasets \cite{yuan2024rrsis,dong2024cross,yang2025largescalereferringremotesensing}, the images are divided into uniform $1080 \times 1080$ patches, thus preserving fine-grained details.
A manual screening step is performed to ensure that the retained tiles exhibit representative characteristics of the LAD imagery.

Following mainstream LAD object detection benchmarks \cite{visdrone,UAVDT,UAV123}, we select eight representative object categories: people, car, motor, bicycle, tricycle, truck, bus and boat.
These categories are commonly encountered and crucial for various drone-based applications.  
To better reflect real-world challenges, we apply a size constraint to ensure that more than $90\%$ of the annotated instances possess a mask coverage ratio below 0.1.
This constraint highlights the prevalence of small and visually inconspicuous objects in LAD scenarios.

\begin{figure}
    \centering
    \includegraphics[width=0.9\linewidth]{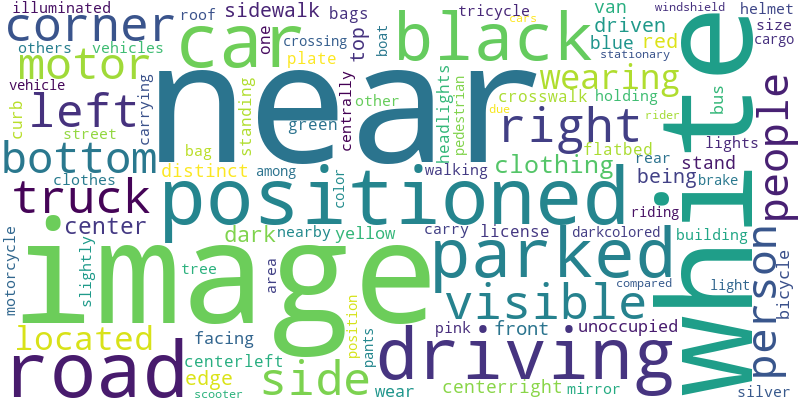}
    \caption{Word cloud for referring expressions in the RIS-LAD dataset.}
    \label{wordcloud}
\end{figure}

\begin{table}[htbp]
  \centering
  \small
    \setlength{\tabcolsep}{1mm}
    \begin{tabular}{c|ccc}
    \toprule
    Dataset & Image Source & \makecell{Shooting\\Angle} & \makecell{Nighttime\\Scene} \\
    \midrule
    RefDIOR      & Google Earth                  & fixed                & \xmark \\
    NWPU-Refer  & Google Earth                  & fixed                & \xmark \\
    RISBench     & Google Earth, GF-2, JL-1      & fixed                & \xmark \\
    RefSegRS     & Helicopter (above 1000m)      & fixed                & \xmark \\
    RRSIS-D     & Google Earth                  & fixed                & \xmark \\
    \midrule
    RIS-LAD     & Drone (30m-100m)           & $30^\circ$–$60^\circ$ & \cmark \\
    \bottomrule
    \end{tabular}
  \caption{Comparison of RIS-LAD and publicly available referring remote sensing image segmentation datasets.}
  \label{tab:dataset_comparison}
\end{table}

\subsubsection{Mask and Expression Generation Pipeline}
Segment Anything Model 2 (SAM-2) \cite{ravi2024sam2} is employed to automatically generate high-quality instance masks, where the oriented bounding boxes from CODrone provide precise prompts to guide the segmentation process.

As shown in Fig.~\ref{annotation pipeline}, referring expressions are generated using the multi-modal large language model (MM-LLM) Qwen2.5-VL \cite{Qwen2.5-VL}.
To capture both instance-level appearance and spatial attributes, each object is cropped and upsampled to improve visual clarity.
The cropped instance and the corresponding full image, marked with a red bounding box that indicates the location, are then fed into Qwen2.5-VL.
This dual-input strategy addresses challenges associated with small object sizes by the MM-LLM’s ability to capture fine-grained visual features and generate accurate spatial descriptions when relying solely on raw coordinates.
As shown in Fig.~\ref{wordcloud}, we further visualize the linguistic diversity through word clouds generated from the refined referring expressions.

\subsubsection{Manual Refinement and Dataset Splitting}
All automatically generated annotations are carefully verified and refined to ensure strong semantic alignment among images, referring expressions, and masks.
This process yields a total of 13,871 high-quality image-text-mask triplets, randomly divided into training, validation, and test sets with a 7: 1: 2 ratio.

\begin{figure*}
    \centering
    \includegraphics[width=0.9\linewidth]{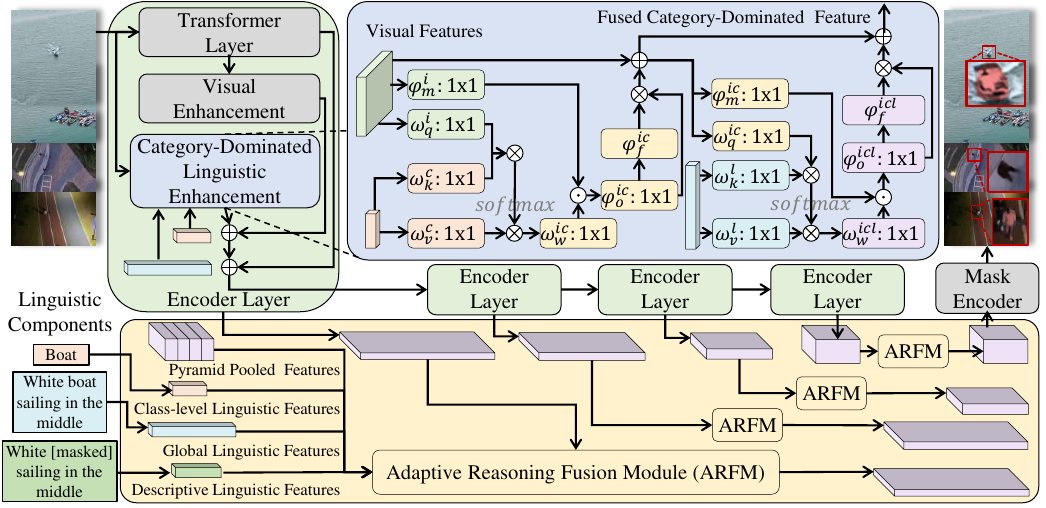}
    \caption{Illustration of the proposed Semantic-Aware Adaptive Reasoning Network (SAARN). The framework comprises two core components: Category-Dominated Linguistic Enhancement (CDLE, top) and Adaptive Reasoning Fusion Module (ARFM, bottom). To mitigate early-stage semantic misalignments, CDLE injects class-level linguistic features at the encoder stage, while ARFM adaptively integrates multi-scale and linguistic features to reason about the correct object in scenes with dense object distributions. $\omega$ and $\varphi$ denote linear and nonlinear mappings, respectively.}
    \label{pipline}
\end{figure*}

\subsubsection{Comparison of RIS-LAD and RRSIS datasets.}
As summarized in Table~\ref{tab:dataset_comparison}, RIS-LAD exhibits distinct characteristics that distinguish it from existing RRSIS datasets \cite{dong2024cross,yang2025largescalereferringremotesensing,yuan2024rrsis,liu2024rotated,lu2025rrsecs}.
First, in contrast to RRSIS datasets that rely on fixed top-down viewpoints, RIS-LAD contains imagery captured from low-altitude drones at oblique angles between $30^\circ$ and $60^\circ$, thus offering greater variation in perspective and object appearance.
Secondly, RIS-LAD uniquely includes nighttime scenes, enabling evaluation under challenging illumination conditions.
Furthermore, it employs a semi-automatic annotation pipeline that integrates MM-LLMs with manual verification, ensuring both annotation quality and efficiency.

\section{Method}
\subsection{Overview}
As shown in Fig.~\ref{pipline}, we propose the Semantic-Aware Adaptive Reasoning Network (SAARN), a framework designed to address category drift and object drift in RLADIS. SAARN incorporates two key components: the Category-Dominated Linguistic Enhancement (CDLE) and the Adaptive Reasoning Fusion Module (ARFM).

Given an input image and its corresponding referring expression, a Swin Transformer \cite{liu2021swin} encoder extracts visual features, and the expression is decomposed into global, class-level, and descriptive components, yielding three types of linguistic features: $l$, $c$, and $d$.
As shown in the bottom-left of Fig.~\ref{pipline}, these linguistic features correspond to the complete expression, the category name, and the descriptive content excluding the category term, respectively. Such disentanglement facilitates targeted modeling of semantic intent, object class information, and detailed spatial or attribute cues, which aligns with the intuitive human reasoning process of first locating the described region or identifying all category-matching objects, and then selecting the target based on specific attributes.

At the encoder stage, CDLE injects class-level guidance through $c$ to align early visual representations with the correct category.
Subsequently, it integrates $l$ to construct category-dominated fused representations, which are then combined via residual fusion with output from the previous layer and the Visual Enhancement module \cite{liu2024rotated} for multi-scale fusion.
In the multi-scale fusion stage, ARFM employs an adaptive integration of pyramid pooled features with $l$, $c$, and $d$ across different scales.
It assigns adaptive weights based on semantic alignment and spatial resolutions, allowing features at each scale to emphasize the most relevant linguistic cues.
The final fused representation is then passed to a mask decoder to generate the segmentation output.

\subsection{Category-Dominated Linguistic Enhancement}
In order to address category drift, we propose the Category-Dominated Linguistic Enhancement (CDLE) module, which selectively injects class-level linguistic features to precisely align early visual representations with the correct categories.
Descriptive linguistic components are deliberately excluded to prevent alignment with visually similar but incorrect objects, thus mitigating category drift.
This module is integrated into each stage of the visual encoder to enable fine-grained cross-modal interaction.

The visual features, or the output from the previous encoder layer, are denoted as $x \in \mathbb{R}^{B \times HW \times D}$, where $B$ is the batch size, $HW$ is the flattened spatial resolution, and $D$ is the dimension of the feature.
The class-level linguistic features $c \in \mathbb{R}^{B \times D_b \times N}$ are derived by extracting the category token from the referring expression and encoding it using a pre-trained BERT model~\cite{devlin2019bert}. $D_b$ and $N$ denote the hidden dimension and the number of linguistic tokens, respectively.
Inspired by \cite{yang2022lavt}, we project both $x$ and $c$ into a shared embedding space to compute the scaled dot-product attention to align category cues and visual features:
\begin{equation}
\alpha^c = \text{softmax}\left( \frac{\omega^i_q(x) \cdot \left(\omega^c_k(c)\right)^\top}{\sqrt{d}} \right) \cdot \omega^c_v(c),
\end{equation}
where each $\omega$ denotes a $1\times1$ convolutional layer applied along the channel dimension, and $d$ is the scaling factor.

Although category-guided attention $\alpha^c$ effectively captures semantic alignment between input and class-level linguistic cues, it tends to over-emphasize category semantics while suppressing other informative visual contexts.
To offset this, a residual gate mechanism is employed, which facilitates adaptive weighting between $\alpha^c$ and the original feature $x$:
\begin{equation}
z^c = \varphi^{ic}_o(\omega^{ic}_w(\alpha^c) \odot \varphi_m^i(x)),
\end{equation}
\begin{equation}
f^{c} = x + z^c \cdot \varphi_{f}^{ic}(z^c),
\end{equation}
where $\varphi_m^i(\cdot)$ and $\varphi^{ic}_o(\cdot)$ denote the convolutional layer followed by GELU. And $\varphi^{ic}_f(\cdot)$ denotes a two-layer linear projection with ReLU \cite{agarap2018deep} and Tanh activations.

Given the global linguistic features $l \in \mathbb{R}^{B \times D_{l} \times N}$ and the category-guided feature $f^{c}$, we again compute the scaled dot-product attention and residual gate,
\begin{equation}
a^l = \text{softmax}\left( \frac{\omega^{ic}_q(f^{c}) \cdot \left(\omega^l_k(l)\right)^\top}{\sqrt{d}} \right) \cdot \omega^l_v(l),
\end{equation}
\begin{equation}
z^l = \varphi_{o}^{icl}(\omega^{icl}_w(\alpha^l) \odot \varphi_m^{ic}(f^{c})),
\end{equation}
\begin{equation}
f^{l} = f^{c} + z^l \cdot \varphi_{f}^{icl}(z^l),
\end{equation}
This design ensures that $l$ is selectively activated in regions corresponding to $c$, thus providing a safeguard mechanism for the fusion between $l$ and $f^c$.
The output $f^l$ retains the discriminative semantics of the referring expression while maintaining a strong alignment with the correct category.
This effectively mitigates attention drift toward visually or semantically similar but incorrect instances.

\begin{figure}[htbp]
    \centering
    \includegraphics[width=0.9\linewidth ]{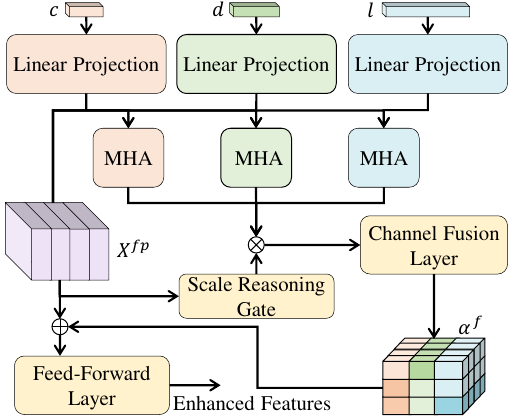}
    \caption{The architecture of the Adaptive Reasoning Fusion Block. Three types of linguistic features are enhanced via Multi-Head Attention (MHA).}
    \label{ARFB}
\end{figure}

\subsection{Adaptive Reasoning Fusion Module}
The Adaptive Reasoning Fusion Module (ARFM) has been proposed to address the problem of object drift induced by densely packed same-category instances in LAD scenarios.
Inspired by the human perception process, in which reasoning typically begins with coarse spatial localization and gradually refines toward specific object attributes, ARFM adopts a progressive fusion strategy.
It establishes a dynamic weighting scheme between the fused multi-scale features $X^{f} = \{x^{f}_{1}, x^{f}_{2}, x^{f}_{3}, x^{f}_{4}\}$ and the linguistic features $l$, $c$, and $d$.
Thus, the model can capture scale-specific semantic cues more effectively, enabling semantic-aware attention across spatial resolutions.
Fine-grained attributes like color and shape are better preserved at higher resolutions, whereas coarse spatial cues such as “top-left corner” become more distinguishable at lower resolutions.

$X^{f}$ is first aligned using pyramid pooling, which is then downsampled to a shared resolution and channel dimension, resulting in $X^{fp} \in \mathbb{R}^{B \times C' \times H' \times W'}$.
These features are then fed into the Adaptive Reasoning Fusion Block (ARFB), which performs cross-modal reasoning across multiple branches, guided by linguistic features.
As shown in Fig.~\ref{ARFB}, ARFB contains three parallel linguistic branches that attend to $c$, $d$ and $l$, respectively.
$X^{fp}$ is projected into a shared embedding space and interacts with each linguistic feature.
Formally, for each linguistic feature $s \in \{l, d, c\}$, we compute:
\begin{align}
q_s &= W^q_sX^{fp}+b^q_s, \quad k_s = W^k_ss+b_s^k, \\
v^s &= W_s^vs+b_s^v, \quad \ \ \ \ \ \alpha_s = \text{MHA}(q_s,k_s,v_s) 
\end{align}
where $W^q_s$, $W^k_s$, and $W^v_s$ are learnable projection matrices for the query, key, and value associated with the linguistic branch $s$, and $\text{MHA}(\cdot)$ denotes the standard multi-head attention \cite{vaswani2017attention}.
The resulting responses $\alpha_s$ encode the cross-modal correlation between $X^{fp}$ and each linguistic feature.

\begin{table*}[!ht]
  \small
  \setlength{\tabcolsep}{1mm}
  \centering
    \begin{tabular}{c|c|cccccccccc|cccc}
    \toprule
    \multirow{2}[2]{*}{Method} & \multirow{2}[2]{*}{Publication} & \multicolumn{2}{c}{P@0.5} & \multicolumn{2}{c}{P@0.6} & \multicolumn{2}{c}{P@0.7} & \multicolumn{2}{c}{P@0.8} & \multicolumn{2}{c|}{P@0.9} & \multicolumn{2}{c}{oIoU} & \multicolumn{2}{c}{mIoU} \\
          &       & Val   & Test  & Val   & Test  & Val   & Test  & Val   & Test  & Val   & Test  & Val   & Test  & Val   & Test \\
    \midrule
    LAVT  & CVPR 2022 & 33.95 & 31.67 & 31.16 & 27.04 & 26.29 & 23.02 & 19.41 & 17.55 & 9.89  & 8.58  & 44.03 & 41.97 & 32.25 & 30.14 \\
    ASDA  & MM 2024 & 35.74 & 33.19 & 29.44 & 26.85 & 23.14 & 20.16 & 13.54 & 12.78 & 4.01  & 3.94  & 38.70 & 37.53 & 35.46 & 33.33 \\
    VATEX & WACV 2025 & 19.27 & 16.07 & 14.76 & 11.87 & 9.74  & 8.47  & 6.02  & 4.67  & 2.08  & 1.63  & 24.83 & 24.27 & 20.32 & 18.53 \\
    \midrule
    LGCE & IEEE TGRS 2024 & 28.22 & 26.38 & 24.07 & 21.82 & 19.99 & 17.55 & 15.11 & 13.32 & 7.88  & 7.06  & 41.72 & 40.75 & 27.68 & 26.17 \\
    FIANet & IEEE TGRS 2024 & 42.91 & 40.21 & 37.61 & 35.22 & 32.59 & 29.93 & 26.43 & 23.71 & 14.83 & 13.21 & 45.24 & 43.39 & 39.61 & 37.44 \\
    RMSIN & CVPR 2024 & 45.85 & 43.36 & 40.97 & 38.11 & 35.24 & 32.36 & 28.01 & 25.08 & 16.33 & 13.75 & 50.17 & 48.82 & 42.08 & 39.60 \\
    RSRefSeg & IGARSS 2025 & 46.35 & 44.73 & 43.48 & 40.21 & 39.76 & 36.30 & 35.17 & 30.91 & 26.58 & 21.46 & 50.04 & 47.71 & 43.42 & 41.16 \\
    CADFormer & JSTARS 2025 & 45.34 & 42.20 & 38.90 & 36.99 & 33.60 & 31.92 & 26.86 & 23.60 & 15.40 & 12.63 & 47.37 & 46.47 & 41.36 & 39.32 \\
    \midrule
    SAARN(\textbf{ours})  &       & 47.06 & 45.02 & 43.12 & 39.34 & 38.25 & 33.37 & 31.81 & 26.24 & 19.27 & 15.31 & \textbf{51.54} & \textbf{49.60} & \textbf{44.30} & \textbf{41.67} \\
    \bottomrule
    \end{tabular}%
        \caption{Comparison with state-of-the-art RIS and RRSIS methods on the proposed RIS-LAD dataset.}
  \label{E1}%
\end{table*}

To adaptively modulate the contribution of each semantic branch, a Scale Reasoning Gate (SRG) module based on $X^{fp}$ is introduced:
\begin{equation}
[w_l, w_d, w_c] = \text{Softmax}(\text{SRG}(X^{fp}))
\end{equation}
$\text{SRG}(\cdot)$ consists of a global average pooling layer followed by two convolutional layers with ReLU activation, which produce three normalized fusion weights.
The weighted attention is computed as,
\begin{equation}
\alpha^f = \text{Fuse}(w_l \alpha_l, \ w_d \alpha_d, \ w_c \alpha_c)
\end{equation}
where $\text{Fuse}(\cdot)$ denotes the channel fusion
layer, which consists of the channel concatenation followed by convolutional projection for compression and alignment.

Finally, $\alpha^f$ is combined with $X^{fp}$ via residual addition and further enhanced through a feed-forward network \cite{vaswani2017attention}, before being passed into the subsequent scale-aware gate fusion module \cite{liu2024rotated}.
ARFM adaptively incorporates different linguistic features into the main semantic path, placing varying emphasis on each feature to enable it to focus on the regions to which it is most sensitive.
This improves context consistency and the quality of object boundary prediction.

\section{Experiments}
\subsection{Implementation Details}
The model is trained on the proposed RIS-LAD dataset for 50 epochs using the AdamW \cite{loshchilovdecoupled} optimizer, with an initial learning rate of $3 \times 10^{-5}$ and a weight decay of 0.01.
A polynomial decay schedule is employed to progressively reduce the learning rate.
All experiments were performed on four NVIDIA RTX 3080 GPUs in a batch size of 8.

For metrics, we report Precision@0.5–0.9 (P@X), Overall Intersection-over-Union (oIoU), and Mean Intersection-over-Union (mIoU).
While P@X highlights accurate predictions, it may overestimate performance on small objects due to their ease of enclosure.
Thus, oIoU and mIoU are used as primary metrics to evaluate segmentation performance.

\subsection{Comparison with SOTA Methods}
As shown in Table~\ref{E1}, the proposed SAARN achieves the best performance across both primary metrics: oIoU and mIoU.
Compared to RMSIN, the state-of-the-art among previous methods \cite{yuan2024rrsis, liu2025cadformer} that does not incorporate pre-trained large vision models (LVMs), our approach increases oIoU from 50.17 to 51.54 on the validation set and further achieves a significant mIoU improvement of 2.07 on the test set.
Notably, SAARN achieves substantial gains under stricter localization thresholds, including +18.0\% in P@0.9 and +13.6\% in P@0.8.

\begin{figure*}[!ht]
    \centering
    \includegraphics[width=1\linewidth]{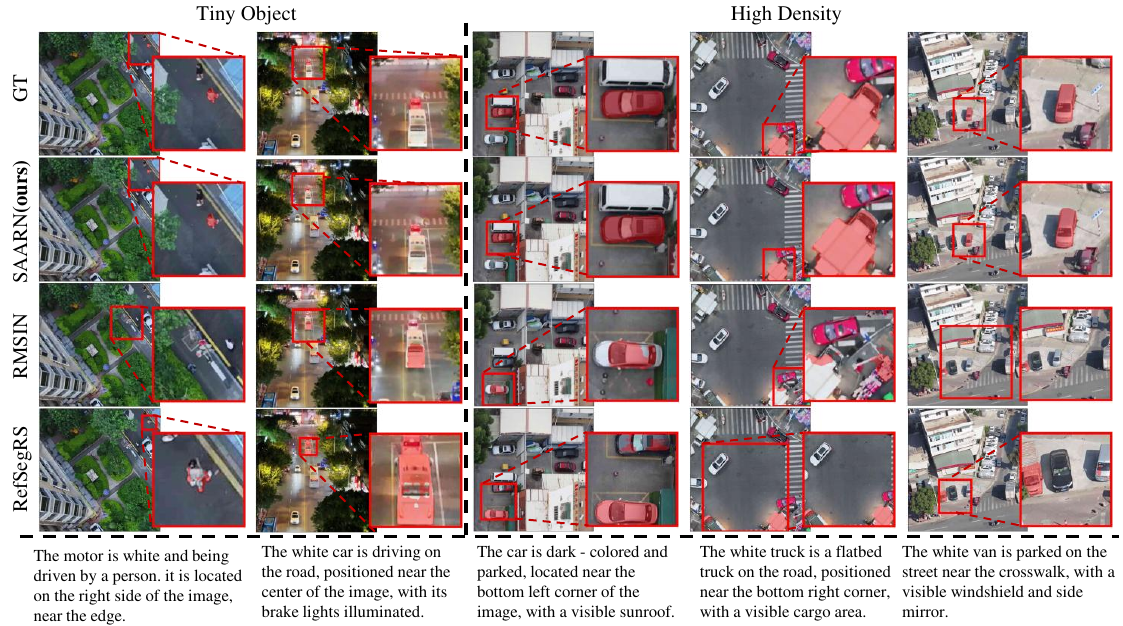}
    \caption{Qualitative comparisons between SAARN and the previous SOTA RMSIN and RefSegRS. The left part illustrates the predictions on tiny-object examples, where category drift easily occurs. The right part shows predictions in high-density scenarios, which are prone to object drift.}
    \label{visual}
\end{figure*}

By incorporating CLIP and SAM, RSRefSeg achieves competitive performance on Precision@0.8–0.9.
However, it still underperforms compared to SAARN on the core segmentation metrics.
Visualization analysis reveals that the inconsistent performance of RSRefSeg due to its tendency to produce over-extended masks, as shown in Fig.~\ref{visual}.
Although these masks may cover the referred object, their boundaries exhibit severe region overgeneralization, leading to lower oIoU and mIoU.
In contrast, SAARN generates more boundary-accurate masks, resulting in stronger segmentation performance.
\begin{table}[htbp]
\small
  \centering
    \begin{tabular}{cc|cccc}
    \toprule
    \multirow{2}[2]{*}{CDLE} & \multirow{2}[2]{*}{ARFM} & \multicolumn{2}{c}{oIoU} & \multicolumn{2}{c}{mIoU} \\
          &       & Val   & Test  & Val   & Test \\
    \midrule
    \xmark     & \xmark     & 49.77 & 48.32 & 42.08 & 39.60 \\
    \xmark     & \cmark     & 51.31 & \textbf{49.70} & 43.97 & 40.68 \\
    \cmark     & \xmark     & 49.82 & 49.28 & 43.31 & 41.02 \\
    \cmark     & \cmark     & \textbf{51.54}  & 49.60 & \textbf{44.30} & \textbf{41.67} \\
    \bottomrule
    \end{tabular}%
  \caption{Ablation study of CDLE and ARFM.}
  \label{tab:ablation_cdle_arfm}%
\end{table}
\begin{table}[htbp]
\small
  \centering
    \begin{tabular}{ccc|cccc}
    \toprule
    \multirow{2}[2]{*}{l} & \multirow{2}[2]{*}{c} & \multirow{2}[2]{*}{d} & \multicolumn{2}{c}{oIoU} & \multicolumn{2}{c}{mIoU} \\
          &       &       & Val   & Test   & Val   & Test \\
    \midrule
    \xmark & \xmark & \xmark & 49.82 & 49.28 & 43.31 & 41.02 \\
    \cmark & \xmark & \xmark & 50.64 & 49.44 & 43.52 & 41.43 \\
    \cmark & \xmark & \cmark & 49.90 & 49.30 & 43.47 & 41.17 \\
    \cmark & \cmark & \xmark & 51.06 & 49.52 & 43.98 & 41.61 \\
    \cmark & \cmark & \cmark & \textbf{51.54} & \textbf{49.60} & \textbf{44.30} & \textbf{41.67} \\
    \bottomrule
    \end{tabular}%
  \caption{Ablation study of linguistic components.}
  \label{ablation_paths}
\end{table}
Compared to FIANet, which also utilizes linguistic decoupling and multi-scale fusion, SAARN performs notably better, owing to its superior instance-level perception in dense distributions of same-class objects.

Conventional RIS methods, such as ASDA and VATEX, face significant performance degradation under the LAD setting, mainly due to the small size of objects and complex background.
In particular, VATEX shows relatively limited performance, partially because it relies on a frozen CLIP image encoder for visual enhancement.
This leads to negative transfer because of the difference in domain between LAD imagery and the CLIP pre-training distribution.
This observation underscores the fundamental domain gap between RLADIS and conventional RIS.

\subsection{Ablation Study}
We conduct ablation studies on the two core modules of SAARN, CDLE and ARFM, to validate their effectiveness.
As shown in Table~\ref{tab:ablation_cdle_arfm}, adding CDLE significantly improves mIoU.
By reinforcing category semantics early in the process, CDLE helps the model focus on the correct object categories, resulting in better instance-level segmentation precision.
In addition, integrating ARFM significantly improves oIoU, increasing it from 49.82 to 51.54 on the validation set.
This improvement results from ARFM’s ability to distinguish densely distributed objects across scales, facilitating more precise localization and significantly reducing false activations.

\subsubsection{Linguistic Components Ablation}
As shown in Table~\ref{ablation_paths}, we further conduct an analysis of the linguistic components used in the ARFM.
Specifically, we evaluate the contributions of the global, class-level and descriptive linguistic feature $l$, $c$ and $d$.
Adding $l$ improves the oIoU on validation set from 49.82 to 50.64, indicating its role in providing global semantic guidance to enhance alignment across multiple scales.
Interestingly, performance drops slightly when $d$ is introduced alongside $l$.
The decrease suggests that, without class-level guidance, the model becomes more susceptible to semantic interference from fine-grained descriptive cues, which may disrupt category-consistent representation learning, especially at coarser resolutions.
In contrast, introducing $c$ significantly improves performance on both oIoU and mIoU.
This highlights the critical role of $c$ in semantic alignment, as it provides strong category-specific constraints.

Finally, combining all three features achieves the best overall performance. This indicates that each linguistic component injected in ARFM plays a distinct but complementary role in the reasoning process, collectively enhancing the model’s ability to accurately localize the referred object.

\subsection{Visualization and Quantitative Results}
We present a qualitative comparison with two of the SOTA RRSIS methods to better illustrate the effectiveness of our proposed model.
As shown in Fig.~\ref{visual}, under typical LAD scenarios, our model accurately identifies the correct object category and effectively distinguishes the object from nearby same-category instances, even in complex backgrounds and under varying drone viewpoints.
In contrast, the compared RRSIS methods exhibit noticeable drift, often leading to misclassification and confusion among closely clustered objects of the same class.
A detailed discussion of model limitations and failure cases is provided in the appendix.

\section{Conclusion}
In this work, we propose RIS-LAD, the first fine-grained dataset designed for RLADIS.
Through detailed analysis, we identified two core challenges specific to this setting: category drift and object drift, both of which significantly degrade the performance of existing RES and RRSIS models.
To address these issues, we propose a semantic-aware adaptive reasoning network which comprises two key modules: CDLE and ARFM.
CDLE enhances early-stage category alignment, while ARFM performs adaptive multi-scale reasoning to mitigate object drift.
The experiments demonstrate the strong performance of our framework and underscore the challenges posed by RIS-LAD, setting a new benchmark for LAD scenarios.

\bibliographystyle{IEEEtran}
\bibliography{example_paper}

\begin{thebibliography}{10}
\providecommand{\url}[1]{#1}
\csname url@samestyle\endcsname
\providecommand{\newblock}{\relax}
\providecommand{\bibinfo}[2]{#2}
\providecommand{\BIBentrySTDinterwordspacing}{\spaceskip=0pt\relax}
\providecommand{\BIBentryALTinterwordstretchfactor}{4}
\providecommand{\BIBentryALTinterwordspacing}{\spaceskip=\fontdimen2\font plus
\BIBentryALTinterwordstretchfactor\fontdimen3\font minus \fontdimen4\font\relax}
\providecommand{\BIBforeignlanguage}[2]{{%
\expandafter\ifx\csname l@#1\endcsname\relax
\typeout{** WARNING: IEEEtran.bst: No hyphenation pattern has been}%
\typeout{** loaded for the language `#1'. Using the pattern for}%
\typeout{** the default language instead.}%
\else
\language=\csname l@#1\endcsname
\fi
#2}}
\providecommand{\BIBdecl}{\relax}
\BIBdecl

\bibitem{banafaa2024comprehensive}
M.~K. Banafaa, {\"O}.~Pepeo{\u{g}}lu, I.~Shayea, A.~Alhammadi, Z.~A. Shamsan, M.~A. Razaz, M.~Alsagabi, and S.~Al-Sowayan, ``A comprehensive survey on 5g-and-beyond networks with uavs: Applications, emerging technologies, regulatory aspects, research trends and challenges,'' \emph{IEEE access}, vol.~12, pp. 7786--7826, 2024.

\bibitem{casanova2025comparison}
J.~J. Casanova, N.~T. Bergmann, J.~E. Kalin, G.~C. Heineck, and I.~C. Burke, ``A comparison of protocols for high-throughput weeds mapping,'' \emph{Smart Agricultural Technology}, p. 101076, 2025.

\bibitem{li2025cooperative}
F.~Li, Z.~Ren, C.~Pan, H.~Ren, J.~Jin, Q.~Wang, and J.~Wang, ``Cooperative sensing and communication beamforming design for low-altitude economy,'' \emph{arXiv preprint arXiv:2506.20244}, 2025.

\bibitem{visdrone}
P.~Zhu, L.~Wen, D.~Du, X.~Bian, H.~Fan, Q.~Hu, and H.~Ling, ``Detection and tracking meet drones challenge,'' \emph{IEEE transactions on pattern analysis and machine intelligence}, vol.~44, no.~11, pp. 7380--7399, 2021.

\bibitem{UAV123}
M.~Mueller, N.~Smith, and B.~Ghanem, ``A benchmark and simulator for uav tracking,'' in \emph{Computer Vision--ECCV 2016: 14th European Conference, Amsterdam, The Netherlands, October 11--14, 2016, Proceedings, Part I 14}.\hskip 1em plus 0.5em minus 0.4em\relax Springer, 2016, pp. 445--461.

\bibitem{DroneCrowd}
L.~Wen, D.~Du, P.~Zhu, Q.~Hu, Q.~Wang, L.~Bo, and S.~Lyu, ``Detection, tracking, and counting meets drones in crowds: A benchmark,'' in \emph{Proceedings of the IEEE/CVF Conference on Computer Vision and Pattern Recognition}, 2021, pp. 7812--7821.

\bibitem{Okutama-Action}
M.~Barekatain, M.~Mart{\'\i}, H.-F. Shih, S.~Murray, K.~Nakayama, Y.~Matsuo, and H.~Prendinger, ``Okutama-action: An aerial view video dataset for concurrent human action detection,'' in \emph{Proceedings of the IEEE conference on computer vision and pattern recognition workshops}, 2017, pp. 28--35.

\bibitem{UAVDT}
D.~Du, Y.~Qi, H.~Yu, Y.~Yang, K.~Duan, G.~Li, W.~Zhang, Q.~Huang, and Q.~Tian, ``The unmanned aerial vehicle benchmark: Object detection and tracking,'' in \emph{Proceedings of the European conference on computer vision (ECCV)}, 2018, pp. 370--386.

\bibitem{sun2025refdrone}
Z.~Sun, Y.~Liu, H.~Zhu, Y.~Gu, Y.~Zou, Z.~Liu, G.-S. Xia, B.~Du, and Y.~Xu, ``Refdrone: A challenging benchmark for referring expression comprehension in drone scenes,'' \emph{arXiv preprint arXiv:2502.00392}, 2025.

\bibitem{li2025aeroreformer}
R.~Li and X.~Zhao, ``Aeroreformer: Aerial referring transformer for uav-based referring image segmentation,'' \emph{arXiv preprint arXiv:2502.16680}, 2025.

\bibitem{liu2025hybrid}
T.~Liu and S.~Li, ``Hybrid global-local representation with augmented spatial guidance for zero-shot referring image segmentation,'' in \emph{Proceedings of the Computer Vision and Pattern Recognition Conference}, 2025, pp. 29\,634--29\,643.

\bibitem{wang2025iterprime}
Y.~Wang, J.~Ni, Y.~Liu, C.~Yuan, and Y.~Tang, ``Iterprime: Zero-shot referring image segmentation with iterative grad-cam refinement and primary word emphasis,'' in \emph{Proceedings of the AAAI Conference on Artificial Intelligence}, no.~8, 2025, pp. 8159--8168.

\bibitem{huang2025densely}
J.~Huang, Z.~Xu, T.~Liu, Y.~Liu, H.~Han, K.~Yuan, and X.~Li, ``Densely connected parameter-efficient tuning for referring image segmentation,'' \emph{arXiv preprint arXiv:2501.08580}, 2025.

\bibitem{chen2025rsrefseg}
K.~Chen, J.~Zhang, C.~Liu, Z.~Zou, and Z.~Shi, ``Rsrefseg: Referring remote sensing image segmentation with foundation models,'' \emph{arXiv preprint arXiv:2501.06809}, 2025.

\bibitem{shi2025multimodal}
L.~Shi and J.~Zhang, ``Multimodal-aware fusion network for referring remote sensing image segmentation,'' \emph{IEEE Geoscience and Remote Sensing Letters}, 2025.

\bibitem{pan2024rethinking}
Y.~Pan, R.~Sun, Y.~Wang, T.~Zhang, and Y.~Zhang, ``Rethinking the implicit optimization paradigm with dual alignments for referring remote sensing image segmentation,'' in \emph{Proceedings of the 32nd ACM International Conference on Multimedia}, 2024, pp. 2031--2040.

\bibitem{lei2024exploring}
S.~Lei, X.~Xiao, T.~Zhang, H.-C. Li, Z.~Shi, and Q.~Zhu, ``Exploring fine-grained image-text alignment for referring remote sensing image segmentation,'' \emph{IEEE Transactions on Geoscience and Remote Sensing}, 2024.

\bibitem{zhang2025referring}
T.~Zhang, Z.~Wen, B.~Kong, K.~Liu, Y.~Zhang, P.~Zhuang, and J.~Li, ``Referring remote sensing image segmentation via bidirectional alignment guided joint prediction,'' \emph{arXiv preprint arXiv:2502.08486}, 2025.

\bibitem{hu2016segmentation}
R.~Hu, M.~Rohrbach, and T.~Darrell, ``Segmentation from natural language expressions,'' in \emph{European conference on computer vision}.\hskip 1em plus 0.5em minus 0.4em\relax Springer, 2016, pp. 108--124.

\bibitem{chng2024mask}
Y.~X. Chng, H.~Zheng, Y.~Han, X.~Qiu, and G.~Huang, ``Mask grounding for referring image segmentation,'' in \emph{Proceedings of the IEEE/CVF Conference on Computer Vision and Pattern Recognition}, 2024, pp. 26\,573--26\,583.

\bibitem{wang2024unveiling}
W.~Wang, T.~Yue, Y.~Zhang, L.~Guo, X.~He, X.~Wang, and J.~Liu, ``Unveiling parts beyond objects: Towards finer-granularity referring expression segmentation,'' in \emph{Proceedings of the IEEE/CVF Conference on Computer Vision and Pattern Recognition}, 2024, pp. 12\,998--13\,008.

\bibitem{shah2024lqmformer}
N.~A. Shah, V.~VS, and V.~M. Patel, ``Lqmformer: Language-aware query mask transformer for referring image segmentation,'' in \emph{Proceedings of the IEEE/CVF Conference on Computer Vision and Pattern Recognition}, 2024, pp. 12\,903--12\,913.

\bibitem{nguyen2025vision}
H.~Nguyen-Truong, E.-R. Nguyen, T.-A. Vu, M.-T. Tran, B.-S. Hua, and S.-K. Yeung, ``Vision-aware text features in referring image segmentation: From object understanding to context understanding,'' in \emph{2025 IEEE/CVF Winter Conference on Applications of Computer Vision (WACV)}.\hskip 1em plus 0.5em minus 0.4em\relax IEEE, 2025, pp. 4988--4998.

\bibitem{yue2024adaptive}
P.~Yue, J.~Lin, S.~Zhang, J.~Hu, Y.~Lu, H.~Niu, H.~Ding, Y.~Zhang, G.~Jiang, L.~Cao \emph{et~al.}, ``Adaptive selection based referring image segmentation,'' in \emph{Proceedings of the 32nd ACM International Conference on Multimedia}, 2024, pp. 1101--1110.

\bibitem{yang2024remamber}
Y.~Yang, C.~Ma, J.~Yao, Z.~Zhong, Y.~Zhang, and Y.~Wang, ``Remamber: Referring image segmentation with mamba twister,'' \emph{European Conference on Computer Vision (ECCV)}, 2024.

\bibitem{chen2024sam4mllm}
Y.-C. Chen, W.-H. Li, C.~Sun, Y.-C.~F. Wang, and C.-S. Chen, ``Sam4mllm: Enhance multi-modal large language model for referring expression segmentation,'' in \emph{European Conference on Computer Vision}.\hskip 1em plus 0.5em minus 0.4em\relax Springer, 2024, pp. 323--340.

\bibitem{Xia_2024_CVPR}
Z.~Xia, D.~Han, Y.~Han, X.~Pan, S.~Song, and G.~Huang, ``Gsva: Generalized segmentation via multimodal large language models,'' in \emph{Proceedings of the IEEE/CVF Conference on Computer Vision and Pattern Recognition (CVPR)}, June 2024, pp. 3858--3869.

\bibitem{sun2022visual}
Y.~Sun, S.~Feng, X.~Li, Y.~Ye, J.~Kang, and X.~Huang, ``Visual grounding in remote sensing images,'' in \emph{Proceedings of the 30th ACM International conference on Multimedia}, 2022, pp. 404--412.

\bibitem{ma2025lscf}
Q.~Ma, L.~Li, X.~Lu, L.~Jiao, F.~Liu, W.~Ma, X.~Liu, and L.~Sun, ``Lscf: Long-term semantic-guidance convformer for referring remote sensing image segmentation,'' \emph{IEEE Transactions on Geoscience and Remote Sensing}, 2025.

\bibitem{lu2025rrsecs}
X.~Lu, L.~Sun, L.~Li, L.~Jiao, Y.~Yang, Z.~Huang, J.~Chai, X.~Liu, F.~Liu, W.~Ma \emph{et~al.}, ``Rrsecs: Referring remote sensing expression comprehension and segmentation,'' \emph{IEEE Geoscience and Remote Sensing Magazine}, 2025.

\bibitem{liu2024rotated}
S.~Liu, Y.~Ma, X.~Zhang, H.~Wang, J.~Ji, X.~Sun, and R.~Ji, ``Rotated multi-scale interaction network for referring remote sensing image segmentation,'' in \emph{Proceedings of the IEEE/CVF Conference on Computer Vision and Pattern Recognition}, 2024, pp. 26\,658--26\,668.

\bibitem{dong2025diffris}
Z.~Dong, Y.~Sun, T.~Liu, and Y.~Gu, ``Diffris: Enhancing referring remote sensing image segmentation with pre-trained text-to-image diffusion models,'' \emph{arXiv preprint arXiv:2506.18946}, 2025.

\bibitem{radford2021learning}
A.~Radford, J.~W. Kim, C.~Hallacy, A.~Ramesh, G.~Goh, S.~Agarwal, G.~Sastry, A.~Askell, P.~Mishkin, J.~Clark \emph{et~al.}, ``Learning transferable visual models from natural language supervision,'' in \emph{International conference on machine learning}.\hskip 1em plus 0.5em minus 0.4em\relax PmLR, 2021, pp. 8748--8763.

\bibitem{kirillov2023segment}
A.~Kirillov, E.~Mintun, N.~Ravi, H.~Mao, C.~Rolland, L.~Gustafson, T.~Xiao, S.~Whitehead, A.~C. Berg, W.-Y. Lo \emph{et~al.}, ``Segment anything,'' in \emph{Proceedings of the IEEE/CVF international conference on computer vision}, 2023, pp. 4015--4026.

\bibitem{dong2024cross}
Z.~Dong, Y.~Sun, T.~Liu, W.~Zuo, and Y.~Gu, ``Cross-modal bidirectional interaction model for referring remote sensing image segmentation,'' \emph{arXiv preprint arXiv:2410.08613}, 2024.

\bibitem{yuan2024rrsis}
Z.~Yuan, L.~Mou, Y.~Hua, and X.~X. Zhu, ``Rrsis: Referring remote sensing image segmentation,'' \emph{IEEE Transactions on Geoscience and Remote Sensing}, 2024.

\bibitem{ye2025clearflexibleprecisecomprehensive}
\BIBentryALTinterwordspacing
K.~Ye, H.~Tang, B.~Liu, P.~Dai, L.~Cao, and R.~Ji, ``More clear, more flexible, more precise: A comprehensive oriented object detection benchmark for uav,'' 2025. [Online]. Available: \url{https://arxiv.org/abs/2504.20032}
\BIBentrySTDinterwordspacing

\bibitem{yang2025largescalereferringremotesensing}
\BIBentryALTinterwordspacing
Z.~Yang, H.~Yao, L.~Tian, X.~Zhao, Q.~Li, and Q.~Wang, ``A large-scale referring remote sensing image segmentation dataset and benchmark,'' 2025. [Online]. Available: \url{https://arxiv.org/abs/2506.03583}
\BIBentrySTDinterwordspacing

\bibitem{ravi2024sam2}
\BIBentryALTinterwordspacing
N.~Ravi, V.~Gabeur, Y.-T. Hu, R.~Hu, C.~Ryali, T.~Ma, H.~Khedr, R.~R{\"a}dle, C.~Rolland, L.~Gustafson, E.~Mintun, J.~Pan, K.~V. Alwala, N.~Carion, C.-Y. Wu, R.~Girshick, P.~Doll{\'a}r, and C.~Feichtenhofer, ``Sam 2: Segment anything in images and videos,'' \emph{arXiv preprint arXiv:2408.00714}, 2024. [Online]. Available: \url{https://arxiv.org/abs/2408.00714}
\BIBentrySTDinterwordspacing

\bibitem{Qwen2.5-VL}
S.~Bai, K.~Chen, X.~Liu, J.~Wang, W.~Ge, S.~Song, K.~Dang, P.~Wang, S.~Wang, J.~Tang, H.~Zhong, Y.~Zhu, M.~Yang, Z.~Li, J.~Wan, P.~Wang, W.~Ding, Z.~Fu, Y.~Xu, J.~Ye, X.~Zhang, T.~Xie, Z.~Cheng, H.~Zhang, Z.~Yang, H.~Xu, and J.~Lin, ``Qwen2.5-vl technical report,'' \emph{arXiv preprint arXiv:2502.13923}, 2025.

\bibitem{liu2021swin}
Z.~Liu, Y.~Lin, Y.~Cao, H.~Hu, Y.~Wei, Z.~Zhang, S.~Lin, and B.~Guo, ``Swin transformer: Hierarchical vision transformer using shifted windows,'' in \emph{Proceedings of the IEEE/CVF international conference on computer vision}, 2021, pp. 10\,012--10\,022.

\bibitem{devlin2019bert}
J.~Devlin, M.-W. Chang, K.~Lee, and K.~Toutanova, ``Bert: Pre-training of deep bidirectional transformers for language understanding,'' in \emph{Proceedings of the 2019 conference of the North American chapter of the association for computational linguistics: human language technologies, volume 1 (long and short papers)}, 2019, pp. 4171--4186.

\bibitem{yang2022lavt}
Z.~Yang, J.~Wang, Y.~Tang, K.~Chen, H.~Zhao, and P.~H. Torr, ``Lavt: Language-aware vision transformer for referring image segmentation,'' in \emph{Proceedings of the IEEE/CVF conference on computer vision and pattern recognition}, 2022, pp. 18\,155--18\,165.

\bibitem{agarap2018deep}
A.~F. Agarap, ``Deep learning using rectified linear units (relu),'' \emph{arXiv preprint arXiv:1803.08375}, 2018.

\bibitem{vaswani2017attention}
A.~Vaswani, N.~Shazeer, N.~Parmar, J.~Uszkoreit, L.~Jones, A.~N. Gomez, {\L}.~Kaiser, and I.~Polosukhin, ``Attention is all you need,'' \emph{Advances in neural information processing systems}, vol.~30, 2017.

\bibitem{loshchilovdecoupled}
I.~Loshchilov and F.~Hutter, ``Decoupled weight decay regularization,'' in \emph{International Conference on Learning Representations}, 2019.

\bibitem{liu2025cadformer}
M.~Liu, X.~Jiang, and X.~Zhang, ``Cadformer: Fine-grained cross-modal alignment and decoding transformer for referring remote sensing image segmentation,'' \emph{IEEE Journal of Selected Topics in Applied Earth Observations and Remote Sensing}, 2025.

\end{thebibliography}






\begin{IEEEbiographynophoto}{Kai Ye}
received the B.S. degree in Artificial Intelligence from Xiamen University, China. He is currently pursuing the Ph.D. degree at the MAC Lab, School of Informatics, Xiamen University, China. His research interests include UAV Perception, open-vocabulary detection and domain incremental learning.
\end{IEEEbiographynophoto}

\begin{IEEEbiographynophoto}{Yingshi Luan}
is currently pursuing the B.S. degree in Computer Science and Technology at the School of Informatics, Xiamen University, China. His research interests include UAV perception and open-vocabulary object detection. 
\end{IEEEbiographynophoto}

\begin{IEEEbiographynophoto}{Zhudi Chen}
is pursuing the B.S. degree at the School of Informatics, Xiamen University, China. His research interests include UAV Perception, open-vocabulary detection, reasoning segmentation, and referring expression segmentation.
\end{IEEEbiographynophoto}

\begin{IEEEbiographynophoto}{Guangyue Meng}
received the B.S. degree in Computer Science and Technology from Ocean University of China, Qingdao, China. He is currently pursuing the M.S. degree at the MAC Lab, School of Informatics, Xiamen University, China. His research focuses on continual learning and open-vocabulary models, with a particular emphasis on their applications in UAV perception.
\end{IEEEbiographynophoto}

\begin{IEEEbiographynophoto}{Pingyang Dai}
 received the M.S. degree in computer science and the Ph.D. degree in automation from
 Xiamen University, Xiamen, China, in 2003 and 2013, respectively.
 
 He is currently a Senior Engineer with the Key Laboratory of Multimedia Trusted Perception and Efficient Computing, Ministry of Education of China, and the School of Informatics, Xiamen University. His research interests include computer vision and machine learning.
\end{IEEEbiographynophoto}

\begin{IEEEbiographynophoto}{Liujuan Cao}
(Member, IEEE) received the B.S., M.S., and Ph.D. degrees from the School of Com
puter Science and Technology, Harbin Engineering University, Harbin, China, in 2005, 2008, and 2013, respectively.

She is currently an Associate Professor with Xiamen University, Xiamen, China. She has authored over 40 papers in the top and major tired journals and conferences, including Conference on Computer Vision and Pattern Recognition (CVPR) and IEEE TRANSACTIONS ON IMAGE PROCESSING (TIP).Her research interests include computer vision and pattern recognition.

Dr. Cao is also the Financial Chair for the IEEE International Workshop on Multimedia Signal Processing (MMSP) 2015, the Workshop Chair for the ACM International Conference on Internet Multimedia Computing and Service (ICIMCS) 2016, and the Local Chair for the Visual and Learning Seminar 2017.
\end{IEEEbiographynophoto}

\vfill

\end{document}